\documentclass[letterpaper, 10 pt, conference]{ieeeconf}  

\IEEEoverridecommandlockouts                              

\usepackage{amsmath,amssymb,amsfonts}
\usepackage[hidelinks]{hyperref}
\usepackage{float}
\usepackage{gensymb}
\usepackage{algorithmic}
\usepackage{graphicx}
\usepackage{textcomp}
\usepackage{url}
\usepackage{tabularx}
\usepackage{cleveref}
\usepackage{xcolor}
\usepackage{siunitx}
\usepackage{bm}
\usepackage{amsmath}
\usepackage{multirow}
\usepackage{ragged2e}
\usepackage{subcaption}

\newcolumntype{L}{>{\raggedright\arraybackslash}X}
\def\BibTeX{{\rm B\kern-.05em{\sc i\kern-.025em b}\kern-.08em
    T\kern-.1667em\lower.7ex\hbox{E}\kern-.125emX}}
\IEEEoverridecommandlockouts                              

\title{\LARGE \bf
Unbiased Active Inference for Classical Control}
\author{Mohamed Baioumy$^{*1}$, Corrado Pezzato$^{*2}$, Riccardo Ferrari$^3$, Nick Hawes$^1$
\thanks{$^*$ Authors with equal contribution}
\thanks{$^{1}$Mohamed Baioumy and Nick Hawes are with the Oxford Robotics Institute, Oxford University,
        {\tt\small [mohamed, nickh]@robots.ox.ax.uk}} %
\thanks{$^{2}$Corrado Pezzato is with the Cognitive Robotics Department, TU Delft,
        {\tt\small c.pezzato@tudelft.nl}} 
\thanks{$^{3}$Riccardo Ferrari is with the Department of Systems and Control, TU Delft,
        {\tt\footnotesize r.ferrari@tudelft.nl}}%
\thanks{This research was partially supported by Ahold Delhaize. All content represents the opinion of the author(s), which is not necessarily shared or endorsed by their respective employers and/or sponsors.}
}

\begin{document}

\maketitle
\thispagestyle{empty}
\pagestyle{empty}

\newcommand{\DriveFullName}{--}
\newcommand{\DriveName}{--}

\begin{abstract}
Active inference is a mathematical framework that originated in computational neuroscience. Recently, it has been demonstrated as a promising approach for constructing goal-driven behavior in robotics. Specifically, the active inference controller (AIC) has been successful on several continuous control and state-estimation tasks. Despite its relative success, some established design choices lead to a number of practical limitations for robot control. These include having a biased estimate of the state, and only an implicit model of control actions. In this paper, we highlight these limitations and propose an extended version of the unbiased active inference controller (u-AIC). The u-AIC maintains all the compelling benefits of the AIC and removes its limitations. Simulation results on a 2-DOF arm and experiments on a real 7-DOF manipulator show the improved performance of the u-AIC with respect to the standard AIC. The code can be found at \url{https://github.com/cpezzato/unbiased_aic}.\\

\end{abstract}

\section{\textbf{Introduction}}
\label{sec:introduction}

Active inference is a mathematical framework prominent in computational neuroscience \cite{friston2017active}. It aims to explain decision-making in biological agents using free-energy minimization. Recent work has led to many schemes based on this framework, for an overview see \cite{lanillos2021survey}. 
Nevertheless, active inference for robot control is still a relatively new field. It is thus of particular importance to understand the limits of such methods. In this paper, we present an analysis of the drawbacks associated with specific design choices within the active inference controller (AIC). Additionally, we propose an extended and improved unbiased AIC (u-AIC), which we initially presented in \cite{baioumy2021ECC} solely for fault-tolerant control. 
\begin{figure}[t!]
    \centering
    \includegraphics[width= 0.75 \linewidth]{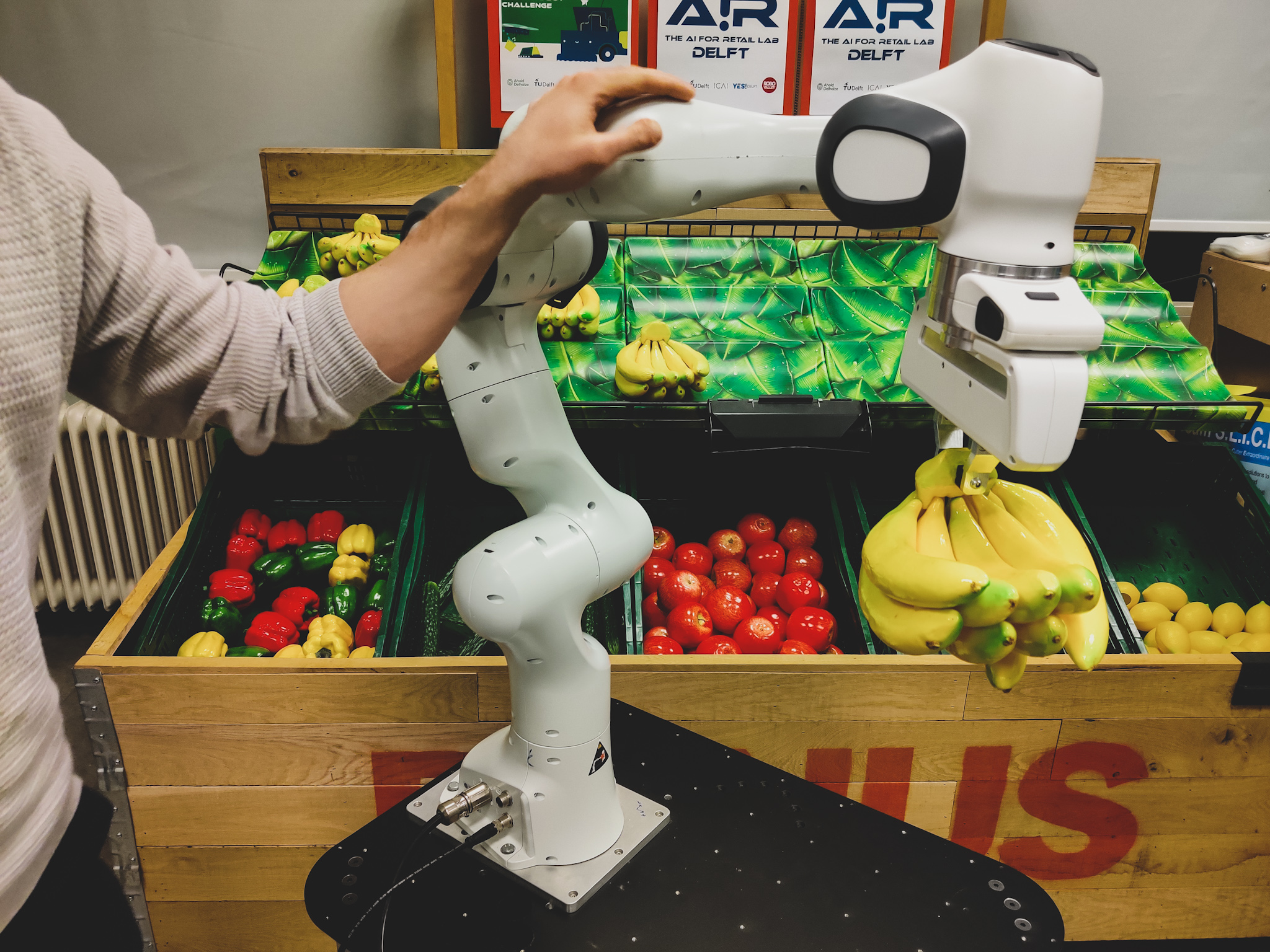}
    \caption{Manipulation of delicate items in a human-shared store environment.}
    \label{fig:example_store}
\end{figure}
The effort in understanding the limits of the AIC for robot control started with our previous work \cite{baioumy2021ECC}, where we specifically analyzed the effect of joint state-estimation and control for fault tolerance. The current paper provides an in-depth analysis of the performance and limits of the AIC more generally (not just fault-tolerant control). Additionally, we formalize the u-AIC. This includes 1) a derivation of the new control architecture and possible extensions, 2) a proof of convergence for both the state estimation and the controlled system, and 3) the application of the u-AIC to a real 7-DOF robot manipulator, which was previously only performed in simulation. 

In this paper, we identify the drawbacks of the AIC and connect them to two root causes. First, in the AIC the estimated state (or \textit{belief}) is biased toward the current goal/target state through a goal prior. This leads to reduced quality in state-estimation \cite{baioumy2020active} but also influences precision learning (learning the precision/covariance of the sensory model) making the model parameters converge to a biased value \cite{baioumy2020active, baltieri2019pid}. A range of issues also arises in fault diagnosis and fault-tolerant control as a result of the goal prior, such as false-positive fault detection \cite{Corrado2020iwai, baioumy2021ECC}. Second, the control action is not explicitly modeled as a random variable in the generative model of the AIC. This causes a range of issues on the control side. In real-world control applications, the integral control law typical of the AIC can cause saturation problems for the actuators when the agent fails to reach a target. This can happen for instance because of a collision. Other limitations from a control perspective are that the AIC does not naturally allow for the incorporation of feed-forward control signals \cite{baioumy2021ECC} which could improve the overall performance of the system. Finally, the motion can be jerky in practice.

Interestingly, all these limitations are present in the AIC but are not intrinsically part of active inference as a general framework. One can design different controllers that stay true to the principles of active inference while mitigating the limitations. To this end, we present the u-AIC, originally introduced in \cite{baioumy2021ECC}.  We demonstrate the properties of the u-AIC in Section \ref{sec:uAIC}, as well as the convergence of both the state estimation and control which is still missing for the AIC \cite{Pezzato2020}.

The \textit{contributions} of this paper are twofold: 1) we concretely demonstrate the limitations of the AIC using both theoretical results and empirical demonstrations. 2) We formalize the u-AIC and show how it overcomes such limitations, providing proof of convergence and extensions to the controller. 
We believe that understanding the properties and the boundaries of the AIC is important for researchers that want to apply it for robot control, such that better and safer systems can be built using this framework. 

\subsection{\textbf{Related work}}
Active inference has been proposed in neuroscience as a general theory of the brain \cite{friston2} and was concisely reported in \cite{buckley} using a notation and language closer to engineering and control. We refer to the active inference controller in \cite{buckley} as AIC, from which the first real-world applications for robot control \cite{Pezzato2020, oliver} were derived. 
Recently, active inference has been applied for robot control on a broader variety of settings and tasks. This includes work on robust state-estimation \cite{Lanillos, meera_colored_noise}, adaptive control \cite{Pezzato2020, baioumy2020active}, fault-tolerant control \cite{baioumy2021ECC, baioumyIWAI2021}, reinforcement learning \cite{tschantz2020reinforcement}, planning under uncertainty \cite{da2020relationship, baioumy2021SSP}, human-robot interaction \cite{chameFrontier2020, Ohata_2020} and more. The recent survey in \cite{lanillos2021survey} provides a detailed overview of active inference for robot control to date. 

Particularly interesting in the context of this paper, is that several recent approaches in robotics have taken inspiration from the free-energy principle and applied it to continuous control and state estimation. Recent works focused on chance-constrained active inference \cite{laar2021chance-constrained}, LQG control \cite{vander_laar_2019_lqg} and model-predictive control \cite{2020ICRA_baioumy}. 

However, while the area of application of active inference in engineering settings is continuously growing, works on in-depth analysis of the AIC from a control theoretic perspective are limited. Work in \cite{baltieri2018probabilistic, baioumy2020active} has shown the relationship between the AIC and PID controllers while the relationship between the AIC, LQR control, and Kalman filters is discussed in \cite{baltieri2020kalman, vander_laar_2019_lqg}. Finally, in \cite{Corrado2020iwai, baioumy2021ECC}, some limitations of the AIC were discussed limited to fault-tolerant control, which motivated the introduction of the u-AIC \cite{baioumy2021ECC}.  
This work focuses on a thorough analysis of the limits of the AIC from a control point of view and proposes an extended version of the u-AIC to address them.

\subsection{\textbf{Structure of the paper}}
The remainder of this paper is organized as follows. After providing the necessary mathematical background in Section~\ref{sec:preliminaries}, Section~\ref{sec:limitations} details the limitations of the AIC when applied to robot control. Section~\ref{sec:uAIC} presents the u-AIC scheme to overcome these limitations. In Section~\ref{sec:relationship} the properties of the u-AIC and its relation with the AIC are highlighted. In Section~\ref{sec:experiments} the theoretical claims are validated in simulation and on a real 7-DOF Franka Emika Panda manipulator. Finally, we draw conclusions in Section~\ref{sec:conclusions}.

\section{\textbf{Preliminaries}}
\label{sec:preliminaries}
In this section, we concisely report the AIC formulation as used in previous work (e.g. \cite{Pezzato2020, baioumy2020active, meo2021multimodal, oliver}), which specified the AIC for robot control according to the theory in \cite{buckley}. In this section, we only report the crucial equations to understand the limitations of the AIC in Section~\ref{sec:limitations}, an interested reader is referred to \cite{Pezzato2020} for all the details. 

\subsection{\textbf{The generative model}}

Active Inference considers an agent in a dynamic environment that receives observations $\bm{y}$ about states $\bm{x}$ at every time-step $t$. The generative model of the agent can then be expressed as:

\begin{equation}
    p(\bm{x}, \bm y) = \underbrace{p(\bm y|\bm{x})}_{observation \hspace{1mm}model \hspace{1mm}} \underbrace{p(\bm{x}).}_{\hspace{1mm} prior} 
    \label{factorization_AIC}
\end{equation}

A visual representation of this model is presented in \cref{fig:factor_graph_demonstrations} (right). The probability distribution $p(\bm y|\bm{x})$ has a mean $\bm g(\bm x)$, which is a mapping from state to observation. The prior $p(\bm{x})$ has a mean $\bm f(\bm x)$, which is a function that encodes the goal state state $\bm \mu_g$. The agent aims to infer the posterior $p(\bm{x}|\bm{y})$ given a model of the agent's world. This means finding the belief $\bm \mu$ over the state $\bm x$ given the observations. This can be achieved by minimizing the so-called variational free energy. If all distributions in eq.~\eqref{factorization_AIC} are Gaussian, the free energy becomes a sum of least square terms \cite{Pezzato2020}. 

\subsection{\textbf{Free-energy}}
The AIC performs joint state estimation and control by minimizing the free-energy $\mathcal{F}$ through a gradient descent scheme. $\mathcal{F}$ is defined as \cite{Pezzato2020}:
\begin{equation}
    \label{eq:generic_F}
	\mathcal{F} (\bm {y}, \bm \mu) = \frac{1}{2}\sum_{i=0}^{n_d-1}\left[ \bm \varepsilon_y^{(i)\top}\Sigma^{-1}_{y^{(i)}}\bm \varepsilon_y^{(i)}+\bm \varepsilon_\mu^{(i)\top}\Sigma^{-1}_{\mu^{(i)}}\bm \varepsilon_\mu^{(i)}\right]+K.
\end{equation}
The free energy is a weighted sum of prediction errors up to a constant $K$ resulting from the derivations \cite{buckley}. The terms $\Sigma^{-1}_{y^{(i)}}$ and $\Sigma^{-1}_{\mu^{(i)}}$ are precision matrices representing the confidence about sensory input and internal beliefs. These can be seen as tuning parameters. The term $n_d$ represents the number of derivatives considered in the control problem. . If we assume as in most cases \cite{Pezzato2020,oliver, baioumy2021ECC} that $n_d = 2$, then state estimation and control are performed on position and velocity. These quantities are internally represented as the beliefs $\bm \mu^{(0)} = \bm \mu$ for positions, and $\bm \mu^{(1)} = \bm \mu'$ for velocities. The terms $\bm \varepsilon_\mu^{(i)}=(\bm{\mu}^{(i+1)}-\bm{f}^{(i)}(\bm \mu))$ and $\bm \varepsilon_y^{(i)}=(\bm{y}^{(i)}-\bm{g}^{(i)}(\bm{\mu}))$ are respectively the \textit{state} and \textit{sensory} prediction errors. 

The function $\bm g(\bm \mu)$ represents the mapping between states and sensory observations. In case of robot control with position and velocity sensors, this is the identity mapping, so $\bm{g}(\bm{\mu}) = \bm{\mu}$.
The function $\bm{f}(\bm{\mu}) = \bm{\mu}_g-\bm{\mu}$ specifies the desired evolution of the dynamics of the system. In this case, \cite{Pezzato2020}, the AIC will make the system behave like a first-order linear system with desired goal position $\bm \mu_g$. By definition \cite{buckley, Pezzato2020}, it holds:
\begin{equation}
\bm{g}^{(i)}=\frac{\partial \bm{g}}{\partial \bm{\mu}}\bm{\mu}^{(i)}, \hspace{1mm} \bm{f}^{(i)}=\frac{\partial \bm{f}}{\partial \bm{\mu}}\bm{\mu}^{(i)}, \hspace{1mm} \bm{g}^{(0)}= \bm g, \hspace{1mm} \bm{f}^{(0)}= \bm f.
\end{equation}

Finally, the terms $\Sigma_{\mu^{(i)}}$ and  $\Sigma_{y^{(i)}}$ are diagonal covariance matrices. In the scalar case, these are simply represented as the variances $\sigma_\mu$ and $\sigma_y$. 

\subsection{\textbf{State estimation}}
State estimation in the AIC is achieved by gradient descent on the free-energy \cite{buckley, friston5, Pezzato2020} with the update rule:
\begin{equation}
    \label{eq:state_update_general}
        \dot{\bm{\tilde{\mu}}} = \frac{d}{dt}\tilde{\bm \mu} -\kappa_\mu\frac{\partial \mathcal{F}}{\partial \bm{\tilde{\mu}}},
\end{equation}
where $\bm{\tilde{\mu}} = [\bm \mu,\ \bm \mu']$, and $t$ refers to time. The term $\kappa_{\mu}$ is a tunable learning rate. 

\subsection{\textbf{Control}}
In the AIC, the control input $\bm u$ is also computed through gradient descent on $\mathcal{F}$. As can be seen by \cref{eq:generic_F}, however, $\mathcal{F}$ does not depend on $\bm u$ explicitly. On the other hand, the actions indirectly influence $\mathcal{F}$ by making the state evolve over time and thus changing the sensory output. We can compute the actions using the chain rule as \cite{buckley, Pezzato2020}:
\begin{equation}
    \label{eq:actions_general}
	\dot{\bm{u}}=-\kappa_a\frac{\partial \bm{\tilde{y}}}{\partial \bm{u}}\frac{\partial \mathcal{F}}{\partial \bm{\tilde{y}}}
\end{equation}
where $\bm{\tilde{y}} = [\bm y,\ \bm y']$, and $\kappa_a$ is the tuning parameter. The term $\frac{\partial \bm{\tilde{y}}}{\partial \bm{u}}$, requires a forward dynamic model and is generally hard to compute in closed-form for non-linear systems. In most previous work \cite{Pezzato2020, oliver, baioumy2021ECC, baioumy2020active} such need has been obviated by introducing a linear approximation. The expression can alternatively be written as a sum for every sensor $\bm y$ in $\bm{\tilde{y}}$. This relationship shows how the control action is directly related to the sensory inputs and the \textit{number} of sensors as seen in the following expression:

\begin{equation}
    \label{eq:actions_general_sum}
	\dot{\bm{u}} \approx -\kappa_a \frac{\partial \mathcal{F}}{\partial \bm{\tilde{y}}} = -\kappa_a \sum_{y} \frac{\partial \mathcal{F}}{\partial \bm{y}}.
\end{equation}


As an example, for a system with a position sensor $\bm{{y}_q}$, and velocity sensor $\bm{y_{\dot{q}}}$ this control law would be:

\begin{equation}
    \dot{\bm{u}}=-\kappa_a\begin{bmatrix}
	\Sigma_{y_q}^{-1}(\bm{y}_q-\bm{\mu})+\Sigma_{y_{\dot{q}}}^{-1}(\bm y_{\dot{q}}-\bm{\mu}')\end{bmatrix}.
	\label{eq:example_control_AIC}
\end{equation}

\section{\textbf{Limitations of the AIC}}
\label{sec:limitations}
In this section, we discuss the limitations of the AIC for robot control. We note that these limitations are not inherent in the active inference framework but rather in how the AIC is constructed. In Section \ref{sec:limit_1} we address \textit{Limitation \#1}: in AIC the belief over the \textit{current} state is biased toward the target the agent aims to reach by means of goal prior. This means that the agent's belief is never accurate, except when the target is reached. In Section \ref{sec:limit_2} we address \textit{Limitation \#2}: the control action is not explicit in the generative model. This means, that one cannot minimize the free energy with respect to the actions directly and instead use the chain rule as in \cref{eq:actions_general}, making assumptions about the linearity of the system. This can cause a saturation problem and does not allow for the incorporation of feed-forward control signals to improve performance. 



\subsection{\textbf{Limitation \#1: biased state estimation}}
\label{sec:limit_1}
\begin{figure}[htb!]
    \centering
    \includegraphics[width = 0.9 \linewidth]{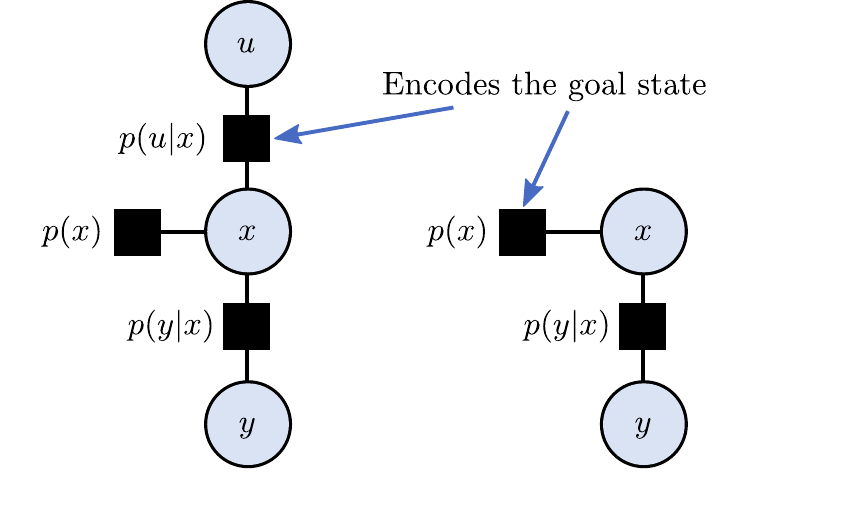}
    \caption{Illustration of the u-AIC (left) and the AIC (right) for a one-dimensional problem. Circles indicate random variables. Black boxes indicate probability distributions.}
    \label{fig:factor_graph_demonstrations}
\end{figure}

Before formally analyzing the controller, we provide a visual explanation using factor graphs. In \cref{fig:factor_graph_demonstrations}, the generative model of the AIC in eq.~\eqref{factorization_AIC} is depicted. Each random variable is represented by a circle, and each probability distribution by a black square. By inspecting this graph, we see that the state $\bm x$ is connected to two distributions. The distribution $p(\bm y| \bm x)$ moves the belief closer to the observed value $\bm y$. This distribution is represented in the free-energy $\mathcal{F}$ by the term $\bm \varepsilon_y^\top\Sigma^{-1}_{y}\bm \varepsilon_y$ where $\bm \varepsilon_y=(\bm{y}-\bm{\mu})$. \textcolor{black}{This is considering, as commonly done, $\bm g(\cdot)$ as the identity mapping},

The second distribution is the prior $p(\bm x)$. This distribution is Gaussian with the function $\bm{f}(\bm{\mu}) = \bm{\mu}_g-\bm{\mu}$ as its mean. The function  encodes the goal state $\bm \mu_g$. This term moves the belief $\bm \mu$ towards the goal state. The belief of the AIC is thus always biased towards the goal state. This is by design. The benefit of this is that now we can also compute a control action that moves the agent to the goal (using \cref{eq:actions_general_sum}). However, the drawback is that state estimation is inaccurate.

\subsubsection{\textbf{Convergence to a biased belief}}
We will now analyze the convergence of the beliefs in the AIC. Let us consider a generic form of the free-energy, as in \cref{eq:generic_F}. For the simplest linear and scalar case with $n_d = 1$ (i.e. one sensor and one actuator, with observable state), this expression can be reduced to:
\begin{equation}
    \mathcal{F} = \frac{1}{2}\left[ \frac{(y-g(\mu))^2}{\sigma_y}+ \frac{(\mu'-f(\mu))^2}{\sigma_\mu} \right] + K
    \label{eq:linear_F_no_sigma}
\end{equation}
Let us consider the generative model of the state dynamics as a first order linear system with unitary time constant,  so $f(\mu) = \mu_g - \mu$, and since the state is observable, so $g(\mu) = \mu$ \cite{Pezzato2020}. At steady state, it holds that
\begin{equation}
    \frac{\partial \mathcal{F}}{\partial {{\mu}}} = \textcolor{black}{-\frac{(y-\mu)}{\sigma_y}+ \frac{(\mu'- \mu_g + \mu)}{\sigma_\mu} = 0}.
\end{equation}
Additionally, when solving for $\mu$ we can show that
\begin{equation}
    \label{eq:mu_convergence}
    \mu=\frac{\sigma_{\mu} y + \sigma_{y} \mu_{g}\textcolor{black}{-\sigma_y\mu'}}{\sigma_{\mu}+\sigma_{y}}.
\end{equation}

From \cref{eq:mu_convergence}, one can notice that the belief $\mu$ is a weighted sum of the sensory measurement and the goal state. The belief is thus always biased towards the goal. We demonstrate incorrect state estimation in simulation (\cref{sec:sim}), which is particularly evident when collisions happen with the environment. The bias in the state estimation leads to two other issues. These are apparent when one tries to optimize model parameters through \textit{precision learning} (or inverse covariance learning), and when applying the AIC for \textit{fault tolerant control}. We expand upon these claims in the following sections.

\subsubsection{\textbf{Biased precision learning}}
Past work on robot control such as \cite{Pezzato2020, oliver} assumed that the precision $\sigma_y^{-1}$ of the observation model is known. Intuitively, this quantifies the confidence about how noisy the sensor is, and can be determined before the deployment of the controller. However, for certain applications where sensory noise is uncertain, one might need to estimate the precision of the sensor online. This is the case in many robotics applications \cite{vega2013cello, pfeifer2017dynamic}. 

There exists a body of work related to active inference which explores precision learning., e.g. methods such as Dynamic Expectation Maximization and generalized filtering \cite{friston2008variational, friston2010generalised}. However, these methods do not perform control. When precision learning is done in conjunction with the active inference controller, we cannot estimate the true precision of the sensor. Work in \cite{baltieri2019pid, baioumy2020active} shows how precision learning in the AIC can be used to find the optimal gains of the controller. In this context, however, the obtained precision of the sensor loses its physical meaning.

As explained earlier, the belief over the current state is biased, and this propagates to the precision of the system. Consider again the simplest linear and scalar case of the free-energy in \cref{eq:linear_F_no_sigma}. The constant $K$ can be expanded to include the variances as \textcolor{black}{(see \cite{buckley})}:
$$K =\textcolor{black}{\frac{1}{2}}\ln \sigma_y \sigma_{\mu}.$$
Considering these terms time-varying, we can take the gradient with respect to the variances as: 
\begin{equation}
    \frac{\partial \mathcal{F}}{\partial \sigma_y} = - \frac{(y-\mu)^2}{2 \sigma_y^2} + \frac{1}{2 \sigma_y}. 
\end{equation}
Setting this expression to zero, we obtain:
\begin{equation}
    \sigma_y = (y - \mu)^2.
\end{equation}

The updated variance depends on $\mu$ which we showed is biased. Thus the precision will also be biased. This is because the agent is estimating the precision as the average square distance between the mean $\mu$ and every measurement $y$. If $\mu$ is biased, then the agent is estimating the variance around a value that is not the true mean. A special case is when the agent already starts at the goal. In that setting, $\mu$ would represent the true mean and the precision would be estimated around the true mean resulting in an accurate precision estimate. Finally, learning the precision of the observation model can be used in the context of control, \textit{but} the obtained precision no longer represents the noise level of the sensor. Instead, it is a combination of the noise level, and how aggressive the controller will act, as shown in \cite{2020ICRA_baioumy}. 

\hspace{1mm}
\subsubsection{\textbf{Limitation in fault-tolerant control}}
The AIC was used in \cite{Corrado2020iwai} for fault tolerant control, where the main idea was that the sensory prediction errors in the free-energy could be used as residuals for fault detection, removing the need for more advanced residual generators. Despite the simplicity of the method for detecting and recovering from broken sensors, the use of prediction errors with biased state estimation can cause several false positives. This is explained in detail in \cite{baioumy2021ECC}, where a first version of the u-AIC has been proposed. We refer the reader to \cite{baioumy2021ECC} for additional details. 
\subsection{\textbf{Limitation \#2: implicit modelling of actions}}
\label{sec:limit_2}
The control law for the AIC is computed using gradient descent on $\mathcal{F}$, as in \cref{eq:actions_general}. Since the action is not explicitly modeled, one resorts to the chain rule. This introduces additional complexity. The term $d\bm y/d\bm\mu$ is generally hard to compute, and researchers approximated it by the identity matrix \cite{baltieri2019pid, Pezzato2020}. Despite the promising results, linearizing the forward dynamics necessarily limits the performance in the presence of non-linear dynamics. Additionally, the control law is driven by sensory prediction errors weighted by the precision, see \cref{eq:example_control_AIC}, and it is de-facto an integrator. This causes three issues in practice.

\textit{First}, the control law in the AIC is directly dependent on the observation. Eq.~\eqref{eq:actions_general_sum} shows that the control law is a sum of sensory prediction errors. This means that, if a system is not observable, it is not controllable. 
More specifically, under the assumptions explained in \cite{baioumy2020active} and \cite{baltieri2019pid}, the active inference controller is equivalent to a PI Controller i.e. PID with $D=0$, a $P$ gain of $\kappa_{u} \Sigma_{y'}^{-1}$ and an $I$ gain of $\kappa_{u} \Sigma_{y}^{-1}$. This assume that the function $\bm{f}(\bm{\mu}) = (\bm{\mu}_g-\bm{\mu})\tau^{-1}$ has a $\tau  \rightarrow 0$. In that case, the belief will approach the goal state $\bm \mu \rightarrow \bm \mu_g$. In \cref{eq:actions_general_sum}, we see that the control law of the AIC is the sum of sensory prediction errors. Every term, in this case, is responsible for an error term in the PID control law. A position sensor is responsible for the $I$ gain, a velocity sensor will result in a $P$ gain, and an acceleration sensor results in a $D$ gain.

\textit{Second}, integral control has a few general drawbacks. For instance, if the goal cannot be reached due to a collision, the magnitude of the control action $\bm u$ will monotonically increase until saturation. This is shown in \cref{sec:experiments} for both simulated and real robots, where we point out that a vanilla implementation of the AIC leads to integrator windup.   

\textit{Third}, the control action cannot be naturally supplemented with a feed-forward signal. In case one has information about the system, designing a feed-forward signal to supplement the feedback controller consistently improves performance. 


\section{\textbf{Unbiased active inference controller}}
\label{sec:uAIC}
A first version of the u-AIC has been introduced in \cite{baioumy2021ECC} in the context of fault-tolerant control. In this section, we generalize, extend, and formally analyze the previously proposed u-AIC for generic control settings, and show how it overcomes the limitations of the AIC. 

\subsection{\textbf{Derivation of the u-AIC}}
In this section, we describe the  u-AIC as introduced in \cite{baioumy2021ECC}, to which an interested reader is referred for more details on the derivations of the following equations.  Let us consider $\bm{x} = [\bm q, \dot{\bm q}]^\top$ and let us define a probabilistic model where actions are modelled explicitly:
\begin{equation}
    p(\bm{x}, \bm{u}, \bm y) = \underbrace{p(\bm{u}|\bm{x})}_{control} \underbrace{p(\bm y|\bm{x})}_{observation \hspace{1 mm}model \hspace{1mm}} \underbrace{p(\bm{x})}_{prior} 
    \label{factorization_u-AIC}
\end{equation}

A visual representation of the u-AIC can be seen in fig.~\ref{fig:factor_graph_demonstrations} (left). Note that with the u-AIC the information about the desired goal to be reached is encoded in the distribution $p(\bm u | \bm x)$. This fundamentally removed the bias of the current state, as will be shown. In this paper, as in \cite{baioumy2020active}, we assume that an accurate dynamic model of the system is not available to keep the solution system agnostic and to highlight once again the adaptability of the controller.

The u-AIC aims at finding the posterior over states as well as the posterior over actions $p(\bm{x}, \bm{u}| \bm y)$. Since the posteriors can be difficult to compute exactly, they are approximated using a variational distribution $Q(\bm{x}, \bm{u})$. We can make use of the mean-field assumption ($Q(\bm{x}, \bm{u}) = Q(\bm{x})Q(\bm{u})$) and the Laplace approximation, and assume the posterior over the state $\bm{x}$ is Gaussian with mean $\bm{{\mu}}_{x}$ \cite{variational}. Similarly for the actions, the posterior $\bm{u}$ is assumed Gaussian with mean $\bm{\mu}_{u}$. By defining the Kullback-Leibler divergence between the variational distribution and the true posterior, one can derive an expression for the free-energy $\mathcal{F}$ as \cite{baioumy2021ECC}:
\begin{equation}
     \mathcal{F} = -\ln p(\bm{\mu}_{u}, \bm{{\mu}}_{x}, \bm y) + C
\end{equation}
Considering \cref{factorization_u-AIC} and assuming Gaussian distributions, $\mathcal{F}$ becomes:
\begin{equation}
\begin{split}
    \label{eq:laplace_F_final_vector}
    \mathcal{F}  
    &=  \frac{1}{2}(\bm{\varepsilon}_{y}^\top \Sigma_{y}^{-1}\bm{\varepsilon}_{y}
    + \bm{\varepsilon}_{x}^\top \Sigma_{x}^{-1}\bm{\varepsilon}_{x}\\
    &+ \bm{\varepsilon}_{u}^\top \Sigma_{u}^{-1}\bm{\varepsilon}_{u} + \ln|\Sigma_{u}\Sigma_{y}\Sigma_{x}|)
    + C,
\end{split}
\end{equation}

The terms $\bm{\varepsilon}_{y_q} = \bm{y}_q-\bm{\mu}$, $\bm{\varepsilon}_{y_{\dot{q}}}=\bm y_{\dot{q}}-\bm{\mu}'$ are the sensory prediction errors respectively for position and velocity sensory inputs. The controller represents the states internally as $\bm \mu_x = [\bm{\mu}, \bm{\mu}']^\top$. The relation between internal state and observation is expressed through the generative model of the sensory input $\bm g = [\bm g_q,\ \bm g_{\dot{q}}]$. Position and velocity encoders directly measure the state, thus $\bm g_q$ and $\bm g_{\dot{q}}$ are linear (identity) mappings. 

Additionally, $\bm{\varepsilon}_{u}$ is the prediction error on the control action while $\bm{\varepsilon}_{x}$ is the prediction error on the state. \textcolor{black}{The latter is computed considering a prediction of the state $\hat{\bm x}$ at the current time-step such that $ \bm{\varepsilon}_{x} = (\bm{\mu}_x - \hat{\bm x} )$. The prediction is a deterministic value $\hat{\bm x} = [\hat{\bm q}, \hat{ \dot{\bm q}}]^\top$ } which can be computed in the same fashion as the prediction step of, for instance, a Kalman filter. The prediction is approximated propagating forward in time the current state belief using the following simplified discrete time model:
\begin{equation}
    \label{eq:euler_integration_a}
    \hat{\bm x}_{k+1} = 
    \begin{bmatrix}
    I & I \Delta t \\
    0 & I
    \end{bmatrix}
    \bm\mu_{x,k}
\end{equation}
where $I$ represents a unitary matrix of suitable size.
This form assumes that the position of each joint is thus computed as the discrete-time integral of the velocity, using a first-order Euler scheme. This approximation can be avoided if a better dynamic model of the system is available, and in that case, predictions can be made using the model itself. Finally, by choosing the distribution $p(\bm u | \bm x)$ to be Gaussian with mean $\bm f^*(\bm{\mu}_x, \bm{\mu}_g)$, we can steer the system towards the target $\bm{\mu}_g$ without biasing the state estimation. This results in  $\bm{\varepsilon}_{u} = (\bm{\mu}_u - \bm f^*(\bm{\mu}_x, \bm{\mu}_g))$. 

In the u-AIC state estimation and control are achieved using gradient descent along the free energy. This leads to:
\begin{equation}
    \label{eq:u-AIC_F_minimize_a}
    \dot{\bm{\mu}}_u = - \kappa_{u}\frac{\partial \mathcal{F}}{\partial \bm{\mu_u}}, \hspace{3 mm}
    \dot{\bm{{\mu}}}_x = - \kappa_{\mu}\frac{\partial \mathcal{F}}{\partial {\bm{{\mu_x}}}},
\end{equation}

\noindent where $\kappa_{u}$ and $\kappa_{\mu}$ are the gradient descent step sizes. The gradient on control can be computed as:
\begin{equation}
    \frac{\partial F}{\partial \bm{\mu}_u}= \Sigma^{-1}_u (\bm{\mu}_u - \bm f^*(\bm{\mu}_x, \bm{\mu}_g))
\end{equation}

\subsection{\textbf{Proof of convergence}}
For the simple 1-dimensional case, the expression of the free energy for the u-AIC can be reduced to:
\begin{equation}
    \mathcal{F} = \frac{1}{2}\left[ \frac{(y- \mu_x )^2}{\sigma_y}+ \frac{(\mu_u - f^{*}(\cdot) )^2}{\sigma_u}+ \frac{(\mu_x - \hat{x})^2}{\sigma_x} \right] +  K
    \label{eq:linear_F_u_aic}
\end{equation}
The belief over the state $\bm \mu_x$ and over the control action $\bm \mu_u$ will converge when: \begin{equation}
        \frac{\partial \mathcal{F}}{\partial \bm{\mu}_u} = 0, \hspace{3 mm}
    \frac{\partial \mathcal{F}}{\partial {\bm{\mu}_x}} = 0,
\end{equation}
\textcolor{black}{At steady state, it holds that $\mu_u = f^*(\mu_x, \mu_g)$. Considering this result, one can show that $\mu_x$ converges to}
\begin{equation}
    \label{eq:mu_convergence_uAIC}
    \mu_x=\frac{\sigma_{x} y + \sigma_{y}\hat{x}} {\sigma_{x}+\sigma_{y}}.
\end{equation}
We can thus see that the control action converges to the function $f^{*}$, which can be chosen for instance as a PID controller. We can also show that belief over the state is a weighted average of the sensory measurement $y$ and the prediction $\hat{x}$. Unlike the AIC, this prediction \textcolor{black}{does not depend on the goal, see eq.~\eqref{eq:mu_convergence}.}

\subsection{\textbf{Extensions of the u-AIC for control}}
The u-AIC can be extended for richer control by modifying the probabilistic model in \cref{factorization_u-AIC}. This is because the control action $\bm u$ is explicitly modeled as a random variable, and thus we can add prior probability distributions to it. This can be done in multiple ways to achieve different objectives.

\subsubsection{\textbf{Adding a feed-forward controller (open-loop)}}

So far the only term depending on the control action now is $p(\bm u| \bm x)$; however, a prior $p(\bm u)$ can be also added. In that case, the generative model would be: 
\begin{equation}
    \frac{1}{\alpha}{p(\bm{u})} {p(\bm{u}|\bm{x})} {p(\bm y|\bm{x})} {p(\bm{x})}
\end{equation}

where $\alpha$ is a normalization constant. Note that, this denominator does not need to be computed explicitly. To achieve state estimation and control, we simply need to minimize the free energy (which maximizes the likelihood of the model). We do not need to exactly compute the free energy or the likelihood. 

This prior can encode a feed-forward signal (open-loop control law). Assuming the prior to be Gaussian with mean $f_{ol}(x)$ and variance $\sigma_{ol}$ (`ol' stands for open-loop), we can show that the control law will converge to:

\begin{equation}
    \mu_u = \frac{f_{ol}(\mu_x)\sigma_u + f^{*}(\mu_g, \mu_x)\sigma_{ol}}{\sigma_{ol}+ \sigma_u}.
    \label{eq:actions_openloop}
\end{equation}

This is the weighted sum of the PID control law (after applying a filter) and the open-loop control law. Note previously, the value of $\sigma_u$ did not matter, now the ratio between $\sigma_u$ and $\sigma_{ol}$ does contribute. Additionally, the expression for $\mathcal{F}$, would contain a quadratic term for the open-loop control law.

\subsubsection{\textbf{Adding control costs}}
Alternative to adding feed-forward control law, we might add a control cost.
This can be achieved by adding the control prior $p_{cc}(\bm{u})$ with a mean of $0$ and variance of $\sigma_{cc}$. This creates a quadratic term in $\mathcal{F}$ of $(\mu_u - 0)^2/\sigma_{cc}$. This is equivalent to a quadratic control cost as seen in classical LQR controllers. 

\subsubsection{\textbf{Smoothing the control action}}
The AIC often exhibits jerky motion. This can be mitigated in the u-AIC by adding a smoothing prior. Let $u$ be a random variable referring to the current control action being executed. We then define $u_p$ as the control action executed at the time previous time-step. A distribution $p(u|u_p)$ can be added to the generative model. This will add a control cost of $(\mu_u - u_p)^2/\sigma_p$ to $\mathcal{F}$. Minimizing this quantity nudges the control action $u$ toward the value of the last control action $u_p$ creating a smoothing effect. This quadratic loss is similar to approaches in Model-predictive control (MPC) \cite{FT_MPC_FDI}. 

\textcolor{black}{
A comparison of the standard AIC and u-AIC against MPC and impedance control can be found in \cite{meo2021multimodal}. Future work could address the comparison of the proposed extensions of the u-AIC against other classical controllers.}

\section{\textbf{Relationship between the AIC and u-AIC}}
\label{sec:relationship}
\subsection{\textbf{Convergence of beliefs}}
The AIC considers the relationship between states and observations without explicitly modeling the control actions. In classical filters, such as the Kalman filter, the prior comes from a prediction step that relies on the previous state and action. In the case of the AIC, the prior essentially predicts the agent to move towards the target. This results in a biased state estimate as seen in \cref{eq:mu_convergence}. In contrast, the u-AIC has an unbiased belief over the state as seen in \cref{eq:mu_convergence_uAIC}. Note that both expressions are almost identical. The AIC converges to the weighted average between the sensory measurement and the goal. The u-AIC on the other hand converges to a weighted average between the sensory measurement and the predicted state (using a model or Euler integration). 
Note that, in the AIC, the \textbf{goal} state is encoded in the prior over the state $p(\bm x)$, while in the u-AIC it is encoded in a separate distribution $p(\bm u|\bm x)$ (see \cref{fig:factor_graph_demonstrations}). 
\subsection{\textbf{Control law}}
In the u-AIC, actions are explicitly modelled and the target state is encoded in the term $p(\bm{u}|\bm{x})$. Explicit actions allow us to directly perform gradient descent on $\mathcal{F}$ with respect to $\bm u$. This is not possible in the general AIC case and the chain rule had to be utilized (see eq.~\eqref{eq:actions_general}). The eventual control law of the AIC is also directly dependent on the measurement and the number of measurements \cref{eq:actions_general_sum}. Thus if part of the system is unobserved, it can not be controlled. The u-AIC on the contrary has a control law irrespective of the number of sensors \cref{eq:mu_convergence_uAIC} and allows us to encode control costs, smoothing effects and feed-forward control signals (see \cref{eq:actions_openloop}).

\subsection{\textbf{General architecture}}
\begin{figure}[htb!]
    \centering
    \includegraphics[width= 0.9 \linewidth]{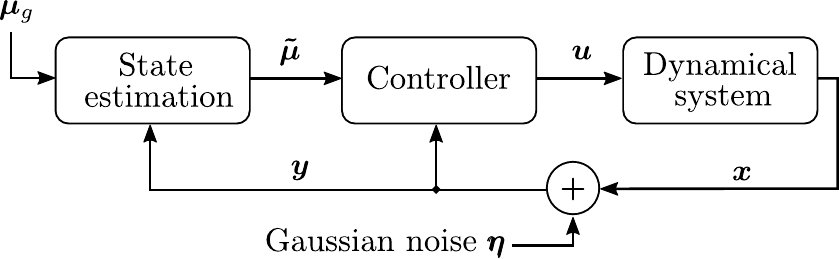}
    \caption{Control diagram of the AIC}
    \label{fig:diagram_AIC}
\end{figure}

\begin{figure}[htb!]
    \centering
    \includegraphics[width= 0.9 \linewidth]{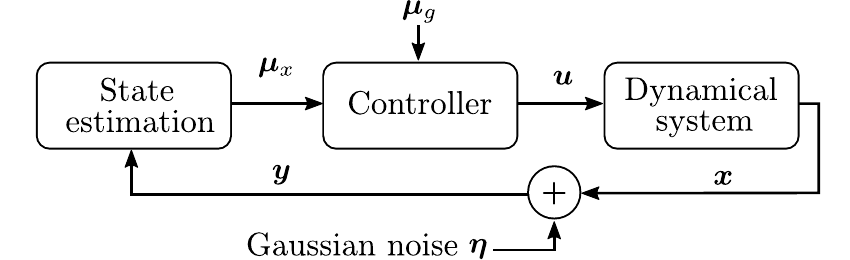}
    \caption{Control diagram of the u-AIC.}
    \label{fig:diagram_uAIC}
\end{figure}

The general architecture (for state estimation and control) for the AIC and u-AIC is also different. In the case of the AIC, state estimation is dependent on the goal state. The goal is encoded in the prior, which biases the state estimation. The controller, on the other hand, is dependent on the (biased) belief and measurements. This is unusual in classical control, see \cref{fig:diagram_AIC}. A discussion on the role of modularity in AIC and the general control architecture can be found in \cite{baltieri2018modularity}.
The u-AIC on the other hand has a more traditional flow of information. State estimation requires the measurements $\bm y$. Then the (unbiased) belief $\bm \mu_x$ and the goal $\bm \mu_g$ are fed into the controller which produces the control action. This is illustrated in \cref{fig:diagram_uAIC}. 


\section{\textbf{Experimental evaluation}}
\label{sec:experiments}

We now showcase the limitations of the AIC explained in Section~\ref{sec:limitations} and compare the performance with the u-AIC, both in simulation and in the real world. 

\subsection{\textbf{Simulation}}
\label{sec:sim}
\begin{figure}[htb!]
    \centering
    \includegraphics[width= 0.90 \linewidth]{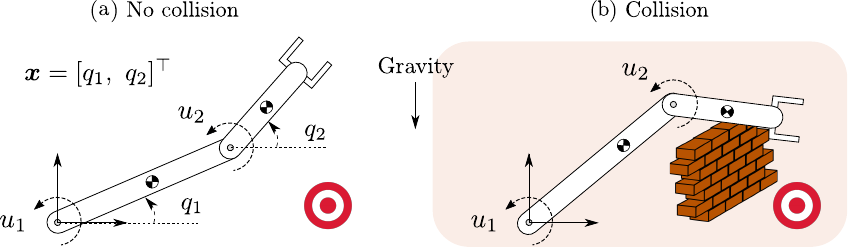}
    \caption{Scenarios considered to illustrate the incorrect state estimation due to the bias towards the target (red dot). In (b) an obstacle occludes the way and blocks the arm.}
    \label{fig:scenarios_bias}
\end{figure}

The simulation scenario is depicted in \cref{fig:scenarios_bias}. The robot has to reach a target in configuration space. During motion, we suppose that at a certain time a collision occurs which prevents the robot from moving further. We assume the collision persists for $t_c = 3s$, then the robot is free to proceed. This allows us to show the incorrect state estimation and the overshoot due to the integral control law of the AIC. In the u-AIC, we observe none of these while maintaining similar performance. 

\begin{figure}[htb!]
    \centering
    \includegraphics[width= 0.99 \linewidth]{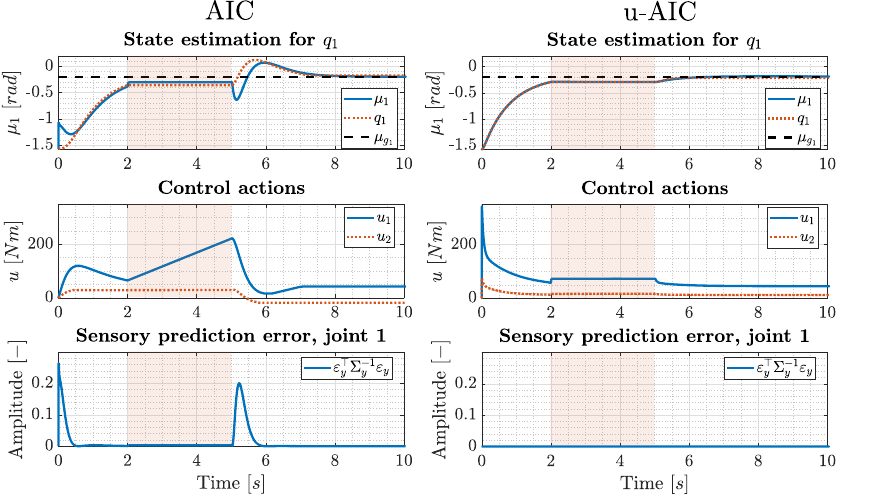}
    \caption{Simulation results of a 2-DOF robot arm collision (orange area) for $t_c = 3s$. For readability, we only report the first joint since the behavior is similar for the second one.}
    \label{fig:limitations_results_2dof}
\end{figure}

The 2-DOF robot arm is equipped with position and velocity sensors $\bm y_q,\ \bm y_{\dot{q}}\in \mathbb{R}^2$ for the two joints affected by zero mean Gaussian noise, as in \cref{fig:scenarios_bias}. We define $\bm y = [\bm y_q,\ \bm y_{\dot{q}}]^\top$. The states $\bm x$ to be controlled are set as the joint positions $\bm q = [q_1,q_2]^\top$ of the robot arm. 
The simulation results for AIC and u-AIC are reported in \cref{fig:limitations_results_2dof}.

The collision scenario in \cref{fig:scenarios_bias} highlights a few limitations:
\subsubsection{\textbf{Incorrect state estimation}}
Let us consider the first row of the plots in \cref{fig:limitations_results_2dof}. When the AIC is not able to reach the desired target due to a collision blocking the path, the belief $\bm \mu$ does not follow the trend of the real state $\bm x$. The belief converges to a value between the sensory reading and the desired goal. For instance, at time $t = 4s$ the sensory reading is $-0.355~[rad]$. Given the current $\mu_g = -0.2~[rad]$ as goal for the first joint, $\sigma_y = \sigma_\mu = 1$, and $\mu'=0.053~[rad/s]$ the belief converges to $-0.304~[rad]$, which is in accordance with \cref{eq:mu_convergence}. On the other hand, the u-AIC converges to the true state. For statistical significance, we ran 100 reaching tasks for each controller, with random blocking collisions of duration between $\in [1, 3]s$ at time $\in [0, 3]s$. Table \ref{tab:sim_stat} reports steady-state error ($e_{ss}$), settling time ($t_s$) after removing the collision, overshoot ($os$), and $RMSE$ between beliefs and joint positions in case of blocking the arm temporarily, averaged over the 100 trials. \begin{table}[h!]
\centering
\begin{tabular}{||c || c c c c||} 
 \hline
 Scenarios  & $e_{ss}\ [rad]$ & $t_s\ [s]$ & $os\ [\%]$ & $RMSE\ [rad]$ \\ [1ex]
 \hline\hline
  AIC & 2.82e-05 & 3.70 & 0.12 & 0.042 \\ 
 u-AIC & 0.0058 & 5.15 & 8.44 & 4.43e-4 \\
 AIC + coll. & 6.07e-05 & 5.04 & 43.90 & 0.1695 \\ 
 u-AIC + coll & 0.0053 & 5.86 & 11.65 & 4.92e-04 \\ 
 \hline
\end{tabular}
\caption{Simulation results of AIC and u-AIC without and with a random collision of random duration, averaged over 100 randomized reaching tasks.}
\label{tab:sim_stat}
\end{table}
As can be seen, while the AIC presents the lowest $e_{ss}$ due to the prominent integration scheme, the $RMSE$ for state estimation is two orders of magnitude higher than for the u-AIC. Incorrect state estimation can also cause several false positives as described in detail in \cite{baioumy2021ECC}. The sensory prediction errors for the AIC can in fact increase also due to collisions with the environment (see the last row of plots in \cref{fig:limitations_results_2dof}). The u-AIC is instead insensitive.

\subsubsection{\textbf{Monotonic increase of control input}}
The second row of the plots in \cref{fig:limitations_results_2dof} displays the control action of one joint. During a collision, the control input computed by the AIC keeps increasing due to the integral nature of the control law employed (see eq.~\eqref{eq:example_control_AIC}). After the collision is removed, the controller necessarily overshoots. 
In the u-AIC, one can saturate only the integral term, and not the entire control law. On average, the AIC has four times the overshoot after a collision with a comparable settling time to the u-AIC.
\subsection{\textbf{7-DOF Panda arm}}
\label{sec:panda}
On the real Panda arm, we compared 1) the reference tracking in configuration space, and 2) the collision behavior in terms of overshoot and commanded control actions resulting from holding the robot arm away from its desired set point. The behaviors of the controllers are best appreciated in the accompanying video\footnote{\url{https://www.youtube.com/watch?v=jI-zX8XvfgI}}.  
\subsubsection{\textbf{Reference tracking}}
\begin{figure}[htb!]
    \centering
    \includegraphics[width= 0.99 \linewidth]{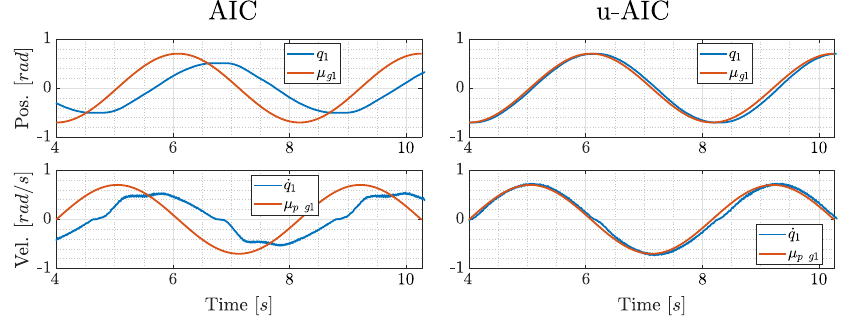}
    \caption{Experiments with real Panda arm on reference tracking. For readability, we only report the first joint.}
    \label{fig:exp_tracking}
\end{figure}

The performance in terms of reference tracking of a sinusoidal wave are reported in \cref{fig:exp_tracking}. \textcolor{black}{The AIC never converges to the reference trajectory, and this is independent of the initial conditions. We highlight this by plotting the response of the two controllers after running them for 4 seconds.}

The AIC performs poorly in terms of tracking errors. This is due to three main reasons: 1) the AIC as defined in \cite{buckley, Pezzato2020} only allows for position reference through its generative model, while the u-AIC can perform position and velocity tracking; 2) a more aggressive tuning of the AIC would result in high overshoots in case of collision, compromising safety. 
\subsubsection{\textbf{Collision behavior}}
\begin{figure}[htb!]
    \centering
    \includegraphics[width= 0.99 \linewidth]{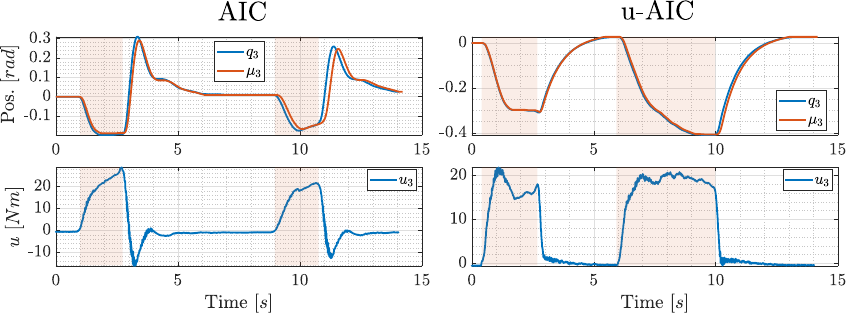}
    \caption{Experiments with real Panda arm during unwanted interaction (orange area). For readability we only report the third joint, being the most affected one. \textcolor{black}{Faster convergence to zero steady-state error for the u-AIC can be achieved through tuning of the integral action.}}
    \label{fig:exp_collision}
\end{figure}

The collision behavior is reported in \cref{fig:exp_collision}. The robot is commanded to keep an initial position while a person is pushing, pulling, and holding the robot. The AIC shows high overshoot which might be dangerous or damage delicate products such as fruits and vegetables in our setting. The u-AIC is instead well-behaved during interaction (see the attached video for a visualization of the results). \textcolor{black}{The same behavior is observed when the robot is disturbed while performing a complicated trajectory.}

\section{Conclusion}
\label{sec:conclusions}

In this paper, we discussed the fundamental limitations of the AIC for robot control and how the u-AIC overcomes them. These limitations arise from the fact that the state estimation is biased towards the goal, and that the control action is not explicitly modeled in the generative model. These cause degraded state estimation and can lead to large overshoots during human-robot interaction. We thoroughly discussed and extended the u-AIC providing a missing proof of convergence. We theoretically demonstrated the limitations of the AIC and provided experimental evidence of how the u-AIC overcomes them, both in simulation and in the real world with a 7-DOF robot manipulator.



\bibliographystyle{IEEEtran}
\bibliography{IEEEabrv,myBib}

\begin{thebibliography}{10}
\providecommand{\url}[1]{#1}
\csname url@rmstyle\endcsname
\providecommand{\newblock}{\relax}
\providecommand{\bibinfo}[2]{#2}
\providecommand\BIBentrySTDinterwordspacing{\spaceskip=0pt\relax}
\providecommand\BIBentryALTinterwordstretchfactor{4}
\providecommand\BIBentryALTinterwordspacing{\spaceskip=\fontdimen2\font plus
\BIBentryALTinterwordstretchfactor\fontdimen3\font minus
  \fontdimen4\font\relax}
\providecommand\BIBforeignlanguage[2]{{%
\expandafter\ifx\csname l@#1\endcsname\relax
\typeout{** WARNING: IEEEtran.bst: No hyphenation pattern has been}%
\typeout{** loaded for the language `#1'. Using the pattern for}%
\typeout{** the default language instead.}%
\else
\language=\csname l@#1\endcsname
\fi
#2}}

\bibitem{friston2017active}
K.~Friston, T.~FitzGerald, F.~Rigoli, P.~Schwartenbeck, and G.~Pezzulo,
  ``Active inference: a process theory,'' \emph{Neural computation}, vol.~29,
  no.~1, pp. 1--49, 2017.

\bibitem{lanillos2021survey}
P.~Lanillos, C.~Meo, C.~Pezzato, A.~A. Meera, M.~Baioumy, W.~Ohata,
  A.~Tschantz, B.~Millidge, M.~Wisse, C.~L. Buckley, \emph{et~al.}, ``Active
  inference in robotics and artificial agents: Survey and challenges,''
  \emph{arXiv preprint arXiv:2112.01871}, 2021.

\bibitem{baioumy2021ECC}
M.~Baioumy, C.~Pezzato, R.~Ferrari, C.~H. Corbato, and N.~Hawes,
  ``Fault-tolerant control of robot manipulators with sensory faults using
  unbiased active inference,'' in \emph{European Control Conf. (ECC)}, 2021.

\bibitem{baioumy2020active}
M.~Baioumy, P.~Duckworth, B.~Lacerda, and N.~Hawes, ``Active inference for
  integrated state-estimation, control, and learning,'' in \emph{Proc of {IEEE}
  Int. Conf. on robotics and automation ({ICRA})}, 2021.

\bibitem{baltieri2019pid}
M.~Baltieri and C.~L. Buckley, ``{PID} control as a process of active inference
  with linear generative models,'' \emph{Entropy}, vol.~21, no.~3, 2019.

\bibitem{Corrado2020iwai}
C.~Pezzato, M.~Baioumy, C.~H. Corbato, N.~Hawes, M.~Wisse, and R.~Ferrari,
  ``Active inference for fault tolerant control of robot manipulators with
  sensory faults,'' in \emph{1st Int. Workshop on Active Inference, ECML PKDD},
  ser. Communications in Computer and Information Science, Springer, Ed., vol.
  1326, 2020.

\bibitem{Pezzato2020}
C.~{Pezzato}, R.~{Ferrari}, and C.~H. {Corbato}, ``A novel adaptive controller
  for robot manipulators based on active inference,'' \emph{IEEE Robotics and
  Automation Letters}, vol.~5, no.~2, pp. 2973--2980, 2020.

\bibitem{friston2}
K.~J. Friston, ``The free-energy principle: a unified brain theory?''
  \emph{Nature Reviews Neuroscience}, vol. 11(2), pp. 27--138, 2010.

\bibitem{buckley}
C.~L. Buckley, C.~S. Kim, S.~McGregor, and A.~K. Seth, ``The free energy
  principle for action and perception: A mathematical review,'' \emph{Journal
  of Mathematical Psychology}, vol.~81, pp. 55--79, 2017.

\bibitem{oliver}
G.~Oliver, P.~Lanillos, and G.~Cheng, ``An empirical study of active inference
  on a humanoid robot,'' \emph{IEEE Transactions on Cognitive and Developmental
  Systems}, 2021.

\bibitem{Lanillos}
P.~Lanillos and G.~Cheng, ``Adaptive robot body learning and estimation through
  predictive coding,'' in \emph{2018 IEEE/RSJ International Conference on
  Intelligent Robots and Systems (IROS)}, 2018.

\bibitem{meera_colored_noise}
A.~Meera and M.~Wisse, ``Free energy principle based state and input observer
  design for linear systems with colored noise,'' in \emph{2020 American
  Control Conf. (ACC)}, 2020, pp. 5052--5058.

\bibitem{baioumyIWAI2021}
M.~Baioumy, C.~Pezzato, C.~H. Corbato, N.~Hawes, and R.~Ferrari, ``Towards
  stochastic fault-tolerant control using precision learning and active
  inference,'' in \emph{IWAI}.\hskip 1em plus 0.5em minus 0.4em\relax Springer,
  2021.

\bibitem{tschantz2020reinforcement}
A.~Tschantz, B.~Millidge, A.~K. Seth, and C.~L. Buckley, ``Reinforcement
  learning through active inference,'' \emph{arXiv preprint arXiv:2002.12636},
  2020.

\bibitem{da2020relationship}
L.~Da~Costa, N.~Sajid, T.~Parr, K.~Friston, and R.~Smith, ``The relationship
  between dynamic programming and active inference: The discrete,
  finite-horizon case,'' \emph{arXiv preprint arXiv:2009.08111}, 2020.

\bibitem{baioumy2021SSP}
M.~Baioumy, P.~Duckworth, B.~Lacerda, and N.~Hawes, ``On solving a stochastic
  shortest-path markov decision process as probabilistic inference,'' in
  \emph{IWAI}.\hskip 1em plus 0.5em minus 0.4em\relax Springer, 2021.

\bibitem{chameFrontier2020}
H.~F. Chame, A.~Ahmadi, and J.~Tani, ``A hybrid human-neurorobotics approach to
  primary intersubjectivity via active inference,'' \emph{Frontiers in
  Psychology}, vol.~11, p. 3207, 2020.

\bibitem{Ohata_2020}
W.~Ohata and J.~Tani, ``Investigation of the sense of agency in social
  cognition, based on frameworks of predictive coding and active inference: A
  simulation study on multimodal imitative interaction,'' \emph{Frontiers in
  Neurorobotics}, vol.~14, Sep 2020.

\bibitem{laar2021chance-constrained}
T.~van~de Laar, {\.I}.~{\c{S}}en{\"o}z, A.~{\"O}z{\c{c}}elikkale, and
  H.~Wymeersch, ``Chance-constrained active inference,'' \emph{Neural
  Computation}, 2021.

\bibitem{vander_laar_2019_lqg}
T.~van~de Laar, A.~{\"O}z{\c{c}}elikkale, and H.~Wymeersch, ``Application of
  the free energy principle to estimation and control,'' \emph{IEEE
  Transactions on Signal Processing}, vol.~69, pp. 4234--4244, 2021.

\bibitem{2020ICRA_baioumy}
M.~Baioumy, M.~Mattamala, and N.~Hawes, ``Variational inference for predictive
  and reactive controllers,'' in \emph{BAIN-PIL workshop, ICRA}, Paris, France,
  2020.

\bibitem{baltieri2018probabilistic}
M.~Baltieri and C.~L. Buckley, ``A probabilistic interpretation of {PID}
  controllers using active inference,'' in \emph{Int. Conf. on Simulation of
  Adaptive Behavior}.\hskip 1em plus 0.5em minus 0.4em\relax Springer, 2018,
  pp. 15--26.

\bibitem{baltieri2020kalman}
------, ``On kalman-bucy filters, linear quadratic control and active
  inference,'' \emph{arXiv preprint arXiv:2005.06269}, 2020.

\bibitem{meo2021multimodal}
C.~Meo and P.~Lanillos, ``Multimodal vae active inference controller,'' in
  \emph{2021 IEEE/RSJ International Conference on Intelligent Robots and
  Systems (IROS)}.\hskip 1em plus 0.5em minus 0.4em\relax IEEE, 2021.

\bibitem{friston5}
K.~Friston, K.~Stephan, B.~Li, and J.~Daunizeau, ``Generalised filtering,''
  \emph{Mathematical Problems in Engineering}, 2010.

\bibitem{vega2013cello}
W.~Vega-Brown and N.~Roy, ``Cello-em: Adaptive sensor models without ground
  truth,'' in \emph{2013 IEEE/RSJ International Conf. on Intelligent Robots and
  Systems}.\hskip 1em plus 0.5em minus 0.4em\relax IEEE, 2013, pp. 1907--1914.

\bibitem{pfeifer2017dynamic}
T.~Pfeifer, S.~Lange, and P.~Protzel, ``Dynamic covariance estimation—a
  parameter free approach to robust sensor fusion,'' in \emph{2017 IEEE
  International Conf. on Multisensor Fusion and Integration for Intelligent
  Systems (MFI)}.\hskip 1em plus 0.5em minus 0.4em\relax IEEE, 2017, pp.
  359--365.

\bibitem{friston2008variational}
K.~J. Friston, N.~Trujillo-Barreto, and J.~Daunizeau, ``Dem: a variational
  treatment of dynamic systems,'' \emph{Neuroimage}, 2008.

\bibitem{friston2010generalised}
K.~Friston, K.~Stephan, B.~Li, and J.~Daunizeau, ``Generalised filtering,''
  \emph{Mathematical Problems in Engineering}, vol. 2010, 2010.

\bibitem{variational}
K.~Friston, J.~Mattout, N.~Trujillo-Barreto, J.~Ashburner, and W.~Penny,
  ``Variational free energy and the {Laplace} approximation,''
  \emph{Neuroimage}, vol. 34(1), pp. 220--234, 2007.

\bibitem{FT_MPC_FDI}
L.~E. Olivier and I.~K. Craig, ``Fault-tolerant nonlinear mpc using particle
  filtering,'' \emph{IFAC-PapersOnLine}, 2016.

\bibitem{baltieri2018modularity}
M.~Baltieri and C.~L. Buckley, ``The modularity of action and perception
  revisited using control theory and active inference,'' \emph{arXiv preprint
  arXiv:1806.02649}, 2018.

\end{thebibliography}

\end{document}